\documentclass{article}

\usepackage{times}
\usepackage{graphicx} 
\usepackage{subfigure} 

\usepackage{natbib}

\usepackage[utf8]{inputenc}
\usepackage{algorithm}
\usepackage{algorithmic}
\usepackage{amssymb}
\usepackage{amsmath}
\usepackage{amsthm}
\usepackage{color}
\usepackage{xcolor}
\usepackage{booktabs}
\usepackage{multirow}
\usepackage{paralist}
\usepackage{titlesec}

\titlespacing{\paragraph}{%
  0pt}{
  0pt}{
  1em}

\usepackage{hyperref}

\newtheorem{theorem}{Theorem}[section]
\newtheorem{definition}[theorem]{Definition}
\newtheorem{lemma}[theorem]{Lemma}
\newtheorem{proposition}[theorem]{Proposition}

\DeclareMathOperator{\bias}{bias}
\DeclareMathOperator{\rank}{rank}

\DeclareMathOperator*{\argmin}{argmin}

\def\E{{\mathbb{E}}}

\newcommand{\user}{u}                       
\newcommand{\users}{U}                      
\newcommand{\movie}{i}                      
\newcommand{\movies}{I}                     

\newcommand{\xpropobs}{X}                   
\newcommand{\xpropobselem}[1]{\xpropobs_{#1}}
\newcommand{\xpropobsum}{\xpropobselem{\user,\movie}}
\newcommand{\xprophid}{X^{hid}}             

\newcommand{\ytrue}{Y}                      
\newcommand{\ytrueelem}[1]{\ytrue_{#1}}
\newcommand{\ytrueum}{\ytrueelem{\user,\movie}}
\newcommand{\ypred}{\hat{Y}}                
\newcommand{\ypredelem}[1]{\ypred_{#1}}
\newcommand{\ypredum}{\ypredelem{\user,\movie}}
\newcommand{\loss}[3]{\delta_{#1}(#2,#3)}   
\newcommand{\lossum}{\loss{\user,\movie}{\ytrue}{\ypred}}
\newcommand{\obs}{O}                        
\newcommand{\obselem}[1]{\obs_{#1}}
\newcommand{\obsum}{\obselem{\user,\movie}}
\newcommand{\probobs}{P}                    
\newcommand{\probobselem}[1]{\probobs_{#1}}
\newcommand{\probobsum}{\probobselem{\user,\movie}}
\newcommand{\estprobobs}{\hat{P}}           
\newcommand{\estprobobselem}[1]{\estprobobs_{#1}}
\newcommand{\estprobobsum}{\estprobobselem{\user,\movie}}
\newcommand{\hypspace}{\mathcal{H}}         
\newcommand{\risk}[1]{R(#1)}                
\newcommand{\risky}{\risk{\ypred}}
\newcommand{\riskemp}[1]{\hat{R}(#1)}                
\newcommand{\riskempy}{\riskemp{\ypred}}
\newcommand{\naiveest}[1]{\hat{R}_{naive}(#1)}     
\newcommand{\naiveesty}{\naiveest{\ypred}}
\newcommand{\ipsest}[2]{\hat{R}_{IPS}(#1|#2)}      
\newcommand{\ipsesty}{\ipsest{\ypred}{\probobs}}
\newcommand{\ipsestyhat}{\ipsest{\ypred}{\estprobobs}}
\newcommand{\snipsest}[2]{\hat{R}_{SNIPS}(#1|#2)}  
\newcommand{\snipsesty}{\snipsest{\ypred}{\probobs}}

\newcommand{\naive}{\emph{Naive}}
\newcommand{\ips}{\emph{IPS}}               
\newcommand{\snips}{\emph{SNIPS}}          
\newcommand{\ipsgold}{\emph{IPS-KNOWN}}     

\newcommand{\ipsnb}{\emph{IPS-NB}}          

\newcommand{\naivemf}{\emph{MF-Naive}}      
\newcommand{\ipsmf}{\emph{MF-IPS}}          
\newcommand{\ipsmfnb}{\emph{MF-IPS-NB}}        

\usepackage[accepted]{icml2016}

\icmltitlerunning{Recommendations as Treatments: Debiasing Learning and Evaluation}

\begin{document} 

\twocolumn[
\icmltitle{Recommendations as Treatments: Debiasing Learning and Evaluation}

\icmlauthor{Tobias Schnabel, Adith Swaminathan, Ashudeep Singh, Navin Chandak, Thorsten Joachims}{}
\icmladdress{Cornell University, Ithaca, NY, USA\hfill {\sc \{tbs49, fa234, as3354, nc475, tj36\}@cornell.edu}}

\icmlkeywords{recommender systems, matrix factorization, propensity, MNAR}

\vskip 0.3in
]

\begin{abstract} 
Most data for evaluating and training recommender systems is subject to selection biases, either through self-selection by the users or through the actions of the recommendation system itself. In this paper, we provide a principled approach to handle selection biases by adapting models and estimation techniques from causal inference. The approach leads to unbiased performance estimators despite biased data, and to a matrix factorization method that provides substantially improved prediction performance on real-world data. We theoretically and empirically characterize the robustness of the approach, and find that it is highly practical and scalable.
\end{abstract} 

\section{Introduction}
Virtually all data for training recommender systems is subject to selection biases. For example, in a movie recommendation system users typically watch and rate those movies that they like, and rarely rate movies that they do not like \cite{pradel2012ranking}. Similarly, when an ad-placement system recommends ads, it shows ads that it believes to be of interest to the user, but will less frequently display other ads. Having observations be conditioned on the effect we would like to optimize (e.g. the star rating, the probability of a click, etc.) leads to data that is Missing Not At Random (MNAR) \cite{rubin2002mnar}. This creates a widely-recognized challenge for 
evaluating recommender systems \cite{Marlin/Zemel/09,de2014reducing}.

We develop an approach to evaluate and train recommender systems that remedies selection biases in a principled, practical, and highly effective way. Viewing recommendation from a causal inference perspective, we argue that exposing a user to an item in a recommendation system is an intervention analogous to exposing a patient to a treatment in a medical study. In both cases, the goal is to accurately estimate the effect of new interventions (e.g. a new treatment policy or a new set of recommendations) despite incomplete and biased data due to self-selection or experimenter-bias. By connecting recommendation to causal inference from experimental and observational data, we derive a principled framework for unbiased evaluation and learning of recommender systems under selection biases.

The main contribution of this paper is four-fold. First, we show how estimating the quality of a recommendation system can be approached with propensity-weighting techniques commonly used in causal inference \cite{Imbens/Rubin/15}, complete-cases analysis \cite{rubin2002mnar}, and other problems \cite{Cortes/etal/08,Bickel/etal/09,Sugiyama/Kawanabe/12}. In particular, we derive unbiased estimators for a wide range of performance measures (e.g. MSE, MAE, DCG). Second, with these estimators in hand, we propose an Empirical Risk Minimization (ERM) framework for learning recommendation systems under selection bias, for which we derive generalization error bounds. Third, we use the ERM framework to derive a matrix factorization method that can account for selection bias while remaining conceptually simple and highly scalable. Fourth, we explore methods to estimate propensities in observational settings where selection bias is due to self-selection by the users, and we characterize the robustness of the framework against mis-specified propensities.

Our conceptual and theoretical contributions are validated in an extensive empirical evaluation. For the task of evaluating recommender systems, we show that our performance estimators can be orders-of-magnitude more accurate than standard estimators commonly used in the past \cite{Bell/etal/07}. For the task of learning recommender systems, we show that our new matrix factorization method substantially outperforms methods that ignore selection bias, as well as existing state-of-the-art methods that perform joint-likelihood inference under MNAR data \cite{HL}. This is especially promising given the conceptual simplicity and scalability of our approach compared to joint-likelihood inference. We provide an implemention of our method, as well as a new benchmark dataset, online\footnote{\href{https://www.cs.cornell.edu/~schnabts/mnar/}{https://www.cs.cornell.edu/$\sim$schnabts/mnar/}}.

\section{Related Work}
Past work that explicitly dealt with the MNAR nature of recommendation data approached the problem as missing-data imputation based on the joint likelihood of the missing data model and the rating model \cite{Marlin07collaborativefiltering,Marlin/Zemel/09,HL}. This has led to sophisticated and highly complex methods. We take a fundamentally different approach that treats both models separately, making our approach modular and scalable.
Furthermore, our approach is robust to mis-specification of the rating model, and we characterize how the overall learning process degrades gracefully under a mis-specified missing-data model. We empirically compare against the state-of-the-art joint likelihood model \cite{HL} in this paper.

Related but different from the problem we consider is recommendation from positive feedback alone \cite{Hu/etal/08,blei2016exposure}. Related to this setting are also alternative approaches to learning with MNAR data \cite{Steck,Steck/11,Lim/etal/15}, which aim to avoid the problem by considering performance measures less affected by selection bias under mild assumptions. Of these works, the approach of \citet{Steck/11} is most closely related to ours, since it defines a recall estimator that uses item popularity as a proxy for propensity. Similar to our work, \citet{Steck,Steck/11} and \citet{Hu/etal/08} also derive weighted matrix factorization methods, but with weighting schemes that are either heuristic or need to be tuned via cross validation. In contrast, our weighted matrix factorization method enjoys rigorous learning guarantees in an ERM framework.

Propensity-based approaches have been widely used in causal inference from observational studies \cite{Imbens/Rubin/15}, as well as in complete-case analysis for missing data \cite{rubin2002mnar,seaman2013review} and in survey sampling \cite{Thompson/12}.  However, their use 
in matrix completion is new to our knowledge. Weighting approaches are also widely used in domain adaptation and covariate shift, where data from one source is used to train for a different problem \citep[e.g.,][]{Huang/etal/06,Bickel/etal/09,Sugiyama/Kawanabe/12}. We will draw upon this work, especially the learning theory of weighting approaches in \cite{Cortes/etal/08,Cortes/etal/10}.

\section{Unbiased Performance Estimation for Recommendation} \label{sec:mnar}
\begin{figure}[t]
\hspace*{-2mm}\includegraphics[width=\textwidth/2]{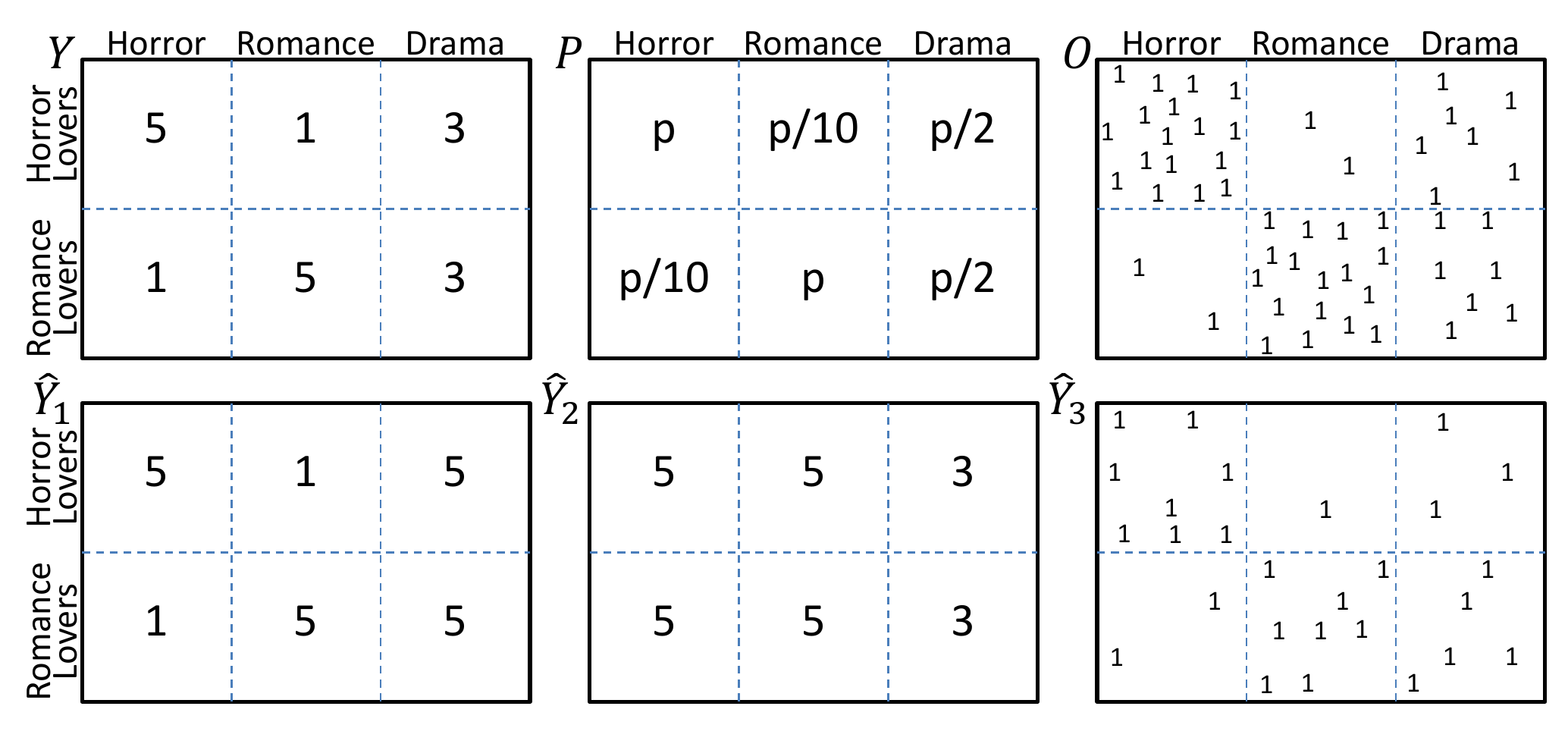}
\vspace*{-10mm}
\caption{Movie-Lovers toy example. Top row: true rating matrix $\ytrue$, propensity matrix $\probobs$, observation indicator matrix $\obs$. Bottom row: two rating prediction matrices $\ypred_1$ and $\ypred_2$, and intervention indicator matrix $\ypred_3$. \label{fig:movielovers}}
\end{figure}
Consider a toy example adapted from \citet{Steck} to illustrate the disastrous effect that selection bias can have on conventional evaluation using a test set of held-out ratings. Denote with $\user \in \{1,...,\users\}$ the users and with $\movie \in \{1,...,\movies\}$ the movies. Figure~\ref{fig:movielovers} shows the matrix of true ratings $\ytrue \in \Re^{\users \times \movies}$ for our toy example, where a subset of users are ``horror lovers'' who rate all horror movies 5 and all romance movies 1. Similarly, there is a subset of ``romance lovers'' who rate just the opposite way. However, both groups rate dramas as 3. 
The binary matrix $\obs \in \{0,1\}^{\users \times \movies}$ in Figure~\ref{fig:movielovers} shows for which movies the users provided their rating to the system, $[\obsum=1] \Leftrightarrow [\ytrueum \mbox{ observed}]$. 
Our toy example shows a strong correlation between liking and rating a movie, and the matrix $\probobs$ describes the marginal probabilities $\probobsum = P(\obsum=1)$ with which each rating is revealed. For this data, consider the following two evaluation tasks.
\subsection{Task 1: Estimating Rating Prediction Accuracy}
For the first task, we want to evaluate how well a predicted rating matrix $\ypred$ reflects the true ratings in $\ytrue$. Standard evaluation measures like Mean Absolute Error (MAE) or Mean Squared Error (MSE) can be written as: 
\begin{eqnarray}
    \risky & = & \frac{1}{\users \!\cdot\! \movies}\sum_{\user=1}^{\users} \sum_{\movie=1}^{\movies} \lossum \enspace , \label{eq:risk} 
\end{eqnarray}
for an appropriately chosen $\lossum$. \vspace{-0.5em}
\begin{eqnarray}
    \mbox{MAE:} && \lossum=|\ytrueum - \ypredum|  \enspace , \label{eq:mae_loss}\\
    \mbox{MSE:} && \lossum=(\ytrueum - \ypredum)^2  \enspace , \label{eq:mse_loss}\\
    \mbox{Accuracy:} && \lossum=\mathbf{1}\{ \ypredum \!=\! \ytrueum\} \enspace .\:\:\:\:\:\:
\end{eqnarray}
Since $\ytrue$ is only partially known, the conventional practice is to estimate $\risky$ using the average over only the observed entries, 
\begin{eqnarray}
    \naiveesty & \!\!\!=\!\!\! & \frac{1}{|\{\!(\user,\movie)\!:\!\obsum \!=\! 1\!\}|}\!\sum_{(\user,\movie):\obsum=1} \!\!\!\!\!\!\!\!\!\lossum \label{eq:naiveest}\enspace .\:\:\:\:\:\:
\end{eqnarray}
We call this the naive estimator, and its naivety leads to a gross misjudgment for the $\ypred_1$ and $\ypred_2$ given in Figure~\ref{fig:movielovers}. Even though $\ypred_1$ is clearly better than $\ypred_2$ by any reasonable measure of performance, $\naiveesty$ will reliably claim that $\ypred_2$ has better MAE than $\ypred_1$. This error is due to selection bias, since 1-star ratings are under-represented in the observed data and $\lossum$ is correlated with $\ytrueum$. More generally, under selection bias, $\naiveesty$ is not an unbiased estimate of the true performance $\risky$ \cite{Steck/13}:
\begin{eqnarray}
    \E_\obs\left[\naiveesty\right] & \not= & \risky \enspace .
\end{eqnarray}
Before we 
design an improved estimator to replace $\naiveesty$, let's turn to a related evaluation task.

\subsection{Task 2: Estimating Recommendation Quality}
Instead of evaluating the accuracy of predicted ratings, we may want to more directly evaluate the quality of a particular recommendation. To this effect, let's redefine $\ypred$ to now encode recommendations as a binary matrix analogous to $\obs$, where $[\ypredum=1] \Leftrightarrow [\movie \mbox{ is recommended to } \user]$,
limited to a budget of $k$ recommendations per user. An example is $\ypred_3$ in Figure~\ref{fig:movielovers}. A reasonable way to measure the quality of a recommendation is the Cumulative Gain (CG) that the user derives from the recommended movies, which we define as the average star-rating of the recommended movies in our toy example\footnote{More realistically, $\ytrue$ would contain quality scores derived from indicators like ``clicked'' and ``watched to the end''.}. CG can again be written in the form of Eq.~\eqref{eq:risk} with
\begin{eqnarray}
\mbox{CG:} && \lossum=(\movies / k) \: \ypredum \cdot \ytrueum  \enspace.
\end{eqnarray}
However, unless users have watched all movies in $\ypred$, we cannot compute CG directly via Eq.~\eqref{eq:risk}. Hence, we are faced with the counterfactual question: how well would our users have enjoyed themselves (in terms of CG), if they had followed our recommendations $\ypred$ instead of watching the movies indicated in $\obs$? Note that rankings of recommendations are similar to the set-based recommendation described above, and measures like Discounted Cumulative Gain (DCG), DCG@k, Precision at k (PREC@k), and others \cite{aslam2006statistical,yilmaz2008simple} also fit in this setting. For those, let the values of $\ypred$ in each row define the predicted ranking, then
\begin{eqnarray}
\mbox{DCG:} && \!\!\!\!\!\!\!\!\!\!\!\lossum\!=\!(\movies / \log(\rank(\ypredum)))\:\ytrueum \enspace,\\
\mbox{PREC@k:} && \!\!\!\!\!\!\!\!\!\!\!\lossum\!=\!(\movies / k)\:\ytrueum\!\cdot\! \mathbf{1}\!\{\rank(\!\ypredum\!) \!\leq\! k\}\enspace .\:\:\:\:\:\:\:
\end{eqnarray}
One approach, similar in spirit to condensed DCG \cite{sakai2007alternatives}, is to again use the naive estimator from Eq.~\eqref{eq:naiveest}. 
However, this and similar estimators are generally biased for $\risky$ \cite{pradel2012ranking,Steck/13}.

To get unbiased estimates of recommendation quality despite missing observations, consider the following connection to estimating average treatment effects of a given policy in causal inference, that was already explored in the contextual bandit setting \cite{Li2011unbiased,dudik11doublyrobust}. If we think of a recommendation as an intervention analogous to treating a patient with a specific drug, in both settings we want to estimate the effect of a new treatment policy (e.g. give drug A to women and drug B to men, or new recommendations $\ypred$). The challenge in both cases is that we have only partial knowledge of how much certain patients (users) benefited from certain treatments (movies) (i.e., $\ytrueum$ with $\obsum=1$), while the vast majority of potential outcomes in $\ytrue$ is unobserved. 

\begin{table*}[tbp]
  \centering
  \scalebox{0.85}{
    \begin{tabular}{rrrrrrrrrr}
\toprule
          & \multicolumn{4}{c}{MAE}       &       & \multicolumn{4}{c}{DCG@50} \\
    \cmidrule{2-5} \cmidrule{7-10}
          & True  &  IPS   & SNIPS & Naive  & \phantom{.}     & True    & IPS   & SNIPS & Naive\\
          \midrule
            REC\_ONES   & 0.102 &    0.102 $\pm$ 0.007   &    0.102 $\pm$ 0.007   &    0.011 $\pm$ 0.001   &       & 30.76  &    30.64 $\pm$ 0.75   &    30.66 $\pm$ 0.74   &    153.07 $\pm$ 2.13  \\
            REC\_FOURS   & 0.026 &    0.026 $\pm$ 0.000   &    0.026 $\pm$ 0.000   &    0.173 $\pm$ 0.001   &       & 52.00 &    51.98 $\pm$ 0.41  &    52.08 $\pm$ 0.58   &    313.48 $\pm$ 2.36   \\
            ROTATE   & 2.579 &    2.581 $\pm$ 0.031   &    2.579 $\pm$ 0.012   &    1.168 $\pm$ 0.003   &       & 12.90  &    13.00 $\pm$ 0.85   &    12.99 $\pm$ 0.83   &    1.38 $\pm$ 0.09  \\
            SKEWED   & 1.306 &    1.304 $\pm$ 0.012   &    1.304 $\pm$ 0.009   &    0.912 $\pm$ 0.002   &       & 24.59 &    24.55 $\pm$ 0.92  &    24.58 $\pm$ 0.93   &    54.87 $\pm$ 1.03  \\
            COARSENED   & 1.320  &    1.314 $\pm$ 0.015   &    1.318 $\pm$ 0.005   &    0.387 $\pm$ 0.002   &       & 46.45 &    46.45 $\pm$ 0.53   &    46.44 $\pm$ 0.70   &    293.27 $\pm$ 1.99  \\
    \bottomrule
    \end{tabular}%
    }
    \vspace*{-2mm}
    \caption{Mean and standard deviation of the Naive, IPS, and SNIPS estimators compared to true MAE and DCG@50 on ML100K. \label{tab:illuexp}}
\end{table*}%

\subsection{Propensity-Scored Performance Estimators}
\label{sec:prop_estimators}
The key to handling selection bias in both of the above-mentioned evaluation tasks lies in understanding the process that generates the observation pattern in $\obs$. This process is typically called the {\em Assignment Mechanism} in causal inference \cite{Imbens/Rubin/15} or the {\em Missing Data Mechanism} in missing data analysis \cite{rubin2002mnar}.
We differentiate the following two settings:
\begin{compactdesc}
    \item[Experimental Setting.] In this setting, the assignment mechanism is under the control of the recommendation system. An example is an ad-placement system that controls which ads to show to which user.
    \item[Observational Setting.] In this setting, the users are part of the assignment mechanism that generates $\obs$. An example is an online streaming service for movies, where users self-select the movies they watch and rate.
\end{compactdesc}
In this paper, we assume that 
the assignment mechanism is probabilistic, meaning that the marginal probability $\probobsum=P(\obsum=1)$ of observing an entry $\ytrueum$ is non-zero for all user/item pairs. This ensures that, in principle, every element of $\ytrue$ could be observed, even though any particular $\obs$ reveals only a small subset.
We refer to $\probobsum$ as the \emph{propensity} of observing $\ytrueum$. 
In the \emph{experimental} setting, we know the matrix $\probobs$ of all propensities, since we have implemented the assignment mechanism. In the \emph{observational} setting, we will need to estimate $\probobs$ from the observed matrix $\obs$. 
We defer the discussion of propensity estimation to Section~\ref{sec:propest}, and focus on the \emph{experimental} setting first. 
\paragraph{IPS Estimator}
The Inverse-Propensity-Scoring (IPS) estimator \cite{Thompson/12,rubin2002mnar,Imbens/Rubin/15}, which applies equally to the task of rating prediction evaluation as to the task of recommendation quality estimation, is defined as,
\begin{eqnarray}
\ipsesty 
         & = & \frac{1}{\users \!\cdot\! \movies}\!\!\!\!\sum_{(\user,\movie):\obsum=1}\!\!\!\! \frac{\lossum}{\probobsum}\label{eq:ips}\enspace.
\end{eqnarray}
Unlike the naive estimator $\naiveesty$, the IPS estimator is unbiased for any probabilistic assignment mechanism.
Note that the IPS estimator only requires the marginal probabilities $\probobsum$ and unbiased-ness is not affected by dependencies within $\obs$:
\vspace{-10mm}
\begin{eqnarray}
\E_\obs\!\left[\ipsesty\right] &\!\!\!\!=\!\!\!\!& 
 \frac{1}{\users \!\cdot\! \movies}\!\sum_\user \sum_\movie \mathbb{E}_{\obsum} \!\!\left[ \frac{\lossum}{\probobsum} \obsum\right]\nonumber \\
&\!\!\!\!=\!\!\!\!& \frac{1}{\users \!\cdot\! \movies}\!\sum_{\user} \sum_{\movie} \lossum = \risky \enspace . \:\:\:\:\:\:\:\:\:\:\label{eq:ipsunbiased} \nonumber
\end{eqnarray}
To characterize the variability of the IPS estimator, however, we assume that observations are independent given $\probobs$, which corresponds to a multivariate Bernoulli model where each $\obsum$ is a biased coin flip with probability $\probobsum$. The following proposition (proof in appendix) provides some intuition about how the accuracy of the IPS estimator changes as the propensities become more ``non-uniform''.

\begin{proposition}[Tail Bound for IPS Estimator]\label{thm:tailbound}
Let $P$ be the independent Bernoulli probabilities of observing each entry. For any given $\ypred$ and $\ytrue$, with probability $1-\eta$, the IPS estimator $\ipsesty$ does not deviate from the true $R(\hat{Y})$ by more than:
\begin{equation}
\left| \ipsesty - \risky \right| \le \frac{1}{\users \!\cdot\! \movies}\sqrt{\frac{\log \frac{2}{\eta}}{2} \sum_{\user,\movie} \rho_{\user,\movie}^2} \enspace , \nonumber
\vspace{-2mm}
\end{equation}
where $\rho_{\user,\movie}=\frac{\lossum}{\probobsum}$ if $\probobsum<1$, and $\rho_{\user,\movie}=0$ otherwise.
\end{proposition}
To illustrate this bound, consider the case of uniform propensities $\probobsum = p$. This means that $n=p\: \users \movies$ elements of $\ytrue$ are revealed in expectation. In this case, the bound is $O(1 / (p \sqrt{\users \movies}) )$. If the $\probobsum$ are non-uniform, the bound can be much larger even if
the expected number of revealed elements, $\sum \probobsum$ is $n$.
We are paying for the unbiased-ness of IPS in terms of variability, and we will evaluate whether this price is well spent throughout the paper.

\paragraph{SNIPS Estimator.} One technique that can reduce variability is the use of control variates \cite{Owen/13}. Applied to the IPS estimator, we know that 
$\E_\obs\left[\sum_{(\user,\movie):\obsum=1} \frac{1}{\probobsum}\right] = \users \cdot \movies$. This yields the Self-Normalized Inverse Propensity Scoring (SNIPS) estimator \cite{Trotter1956,Swaminathan/Joachims/15d}
\begin{equation}
\snipsesty = \frac{\sum_{(\user,\movie):\obsum=1} \frac{\lossum}{\probobsum}}{\sum_{(\user,\movie):\obsum=1} \frac{1}{\probobsum}}\label{eq:snips}\enspace .
\end{equation}
The SNIPS estimator often has lower variance than the IPS estimator but has a small bias \cite{Hesterberg1995}.

\subsection{Empirical Illustration of Estimators} \label{sec:evalillu}
To illustrate the effectiveness of the proposed estimators 
we conducted an experiment on the semi-synthetic ML100K dataset described  in Section~\ref{sec:expbiaseval}. For this dataset, $\ytrue$ is completely known so that we can compute true performance via Eq.~\eqref{eq:risk}. The probability $\probobsum$ of observing a rating $\ytrueum$ was chosen to mimic the observed marginal rating distribution in the original ML100K dataset (see Section~\ref{sec:expbiaseval}) such that, on average, 5\% of the $\ytrue$ matrix was revealed. 

Table~\ref{tab:illuexp} shows the results for estimating rating prediction accuracy via MAE and recommendation quality via DCG@50 for the following five prediction matrices $\ypred_i$. Let $|\ytrue=r|$ be the number of $r$-star ratings in $\ytrue$.
\begin{compactdesc}
    \item[REC\underline{{ }{ }}ONES:] The prediction matrix $\ypred$ is identical to the true rating matrix $\ytrue$, except that $|\{(\user,\movie):\ytrueum=5\}|$ randomly selected true ratings of $1$ are flipped to $5$. This means half of the predicted fives are true fives, and half are true ones. 
    \item[REC\underline{{ }{ }}FOURS:] Same as REC\underline{{ }{ }}ONES, but flipping $4$-star ratings instead. 
    \item[ROTATE:] For each predicted rating $\ypredum = \ytrueum - 1$ when $\ytrueum \ge 2$, and $\ypredum = 5$ when $\ytrueum = 1$.
    \item[SKEWED:] Predictions $\ypredum$ are sampled from $\mathcal{N}(\ypredum^{raw}|\mu=\ytrueum,\sigma=\frac{6-\ytrueum}{2})$ and clipped to the interval $[0,6]$.
    \item[COARSENED:] If the true rating $\ytrueum \le 3$, then $\ypredum=3$. Otherwise $\ypredum=4$.
\end{compactdesc}
Rankings for DCG@50 were created by sorting items according to $\ypred_i$ for each user.
In Table~\ref{tab:illuexp}, we report the average and standard deviation of estimates over 50 samples of $\obs$ from $\probobs$.
We see that the mean IPS estimate perfectly matches the true performance for both MAE and DCG as expected.
The bias of SNIPS is negligible as well. The naive estimator is severely biased and its estimated MAE incorrectly ranks the prediction matrices $\ypred_i$ (e.g. it ranks the performance of \emph{REC\underline{{ }{ }}ONES} higher than \emph{REC\underline{{ }{ }}FOURS}). The standard deviation of IPS and SNIPS is substantially smaller than the bias that Naive incurs. Furthermore, SNIPS manages to reduce the standard deviation of IPS for MAE but not for DCG. We will empirically study these estimators more comprehensively in Section~\ref{sec:exp}. 

\section{Propensity-Scored Recommendation Learning}
We will now use the unbiased estimators from the previous section in an Empirical Risk Minimization (ERM) framework for learning, prove generalization error bounds, and derive a matrix factorization method for rating prediction.

\subsection{ERM for Recommendation with Propensities}
\label{sec:erm_prop}
Empirical Risk Minimization underlies many successful learning algorithms like SVMs \cite{Cortes/Vapnik/95a}, Boosting \cite{schapire}, and Deep Networks \cite{bengio2009deep}. Weighted ERM approaches have been effective for cost-sensitive classification, domain adaptation and co-variate shift \cite{Zadrozny/etal/03, Bickel/etal/09,Sugiyama/Kawanabe/12}. We adapt ERM to our setting by realizing that Eq.~\eqref{eq:risk} corresponds to an expected loss (i.e.\ risk) over the data generating process $P(\obs|\probobs)$.
Given a sample from $P(\obs|\probobs)$, we can think of the IPS estimator from Eq.~\eqref{eq:ips} as the Empirical Risk $\riskempy$ that estimates $\risky$ for any $\ypred$.
\begin{definition}[Propensity-Scored ERM for Recommendation]
Given training observations $\obs$ from $\ytrue$ with marginal propensities $\probobs$, given a hypothesis space $\hypspace$ of predictions $\ypred$, and given a loss function $\lossum$, ERM selects the $\ypred \in \hypspace$ that optimizes:
\vspace{-0.7em}
\begin{eqnarray}
     \ypred^{ERM} & = & \argmin_{\ypred \in \hypspace} \left\{ \ipsesty \right\} \enspace .
\end{eqnarray}
\end{definition}
\vspace{-1.2em}
Using the \snips\ estimator does not change the argmax. To illustrate the validity of the propensity-scored ERM approach, we state the following generalization error bound (proof in appendix) similar to \citet{Cortes/etal/10}. We consider only finite $\hypspace$ for the sake of conciseness.
\begin{theorem}[Propensity-Scored ERM Generalization Error Bound]
\label{thm:prop_erm}
For any finite hypothesis space of predictions $\hypspace=\{\ypred_1, ..., \ypred_{|\hypspace|}\}$ and loss $0 \le \lossum \le \Delta$, the true risk $\risky$ of the empirical risk minimizer $\ypred^{ERM}$ from $\hypspace$ using the IPS estimator, given training observations $\obs$ from $\ytrue$ with independent Bernoulli propensities $\probobs$, is bounded with probability $1-\eta$ by:
\begin{eqnarray}
\risk{\ypred^{ERM}} & \!\le\! & \ipsest{\ypred^{ERM}}{\probobs} + \nonumber \\
&& \frac{\Delta}{\users \!\cdot\! \movies} \sqrt{\frac{\log\left(2 |\hypspace| / \eta\right)}{2}} \sqrt{\sum_{\user,\movie} \frac{1}{\probobsum^2}}\enspace . \:\:\:\:\:\:\: \label{eq:ermbound}
\end{eqnarray}
\end{theorem}

\subsection{Propensity-Scored Matrix Factorization}
We now use propensity-scored ERM to derive a matrix factorization method for the problem of rating prediction. Assume a standard rank-$d$-restricted and $L_2$-regularized matrix factorization model $\ypredum=v_\user^T w_\movie + a_{\user} + b_{\movie} + c$ with user, item, and global offsets as our hypothesis space $\hypspace$. Under this model, propensity-scored ERM leads to the following training objective:
\begin{eqnarray}
\!\!\!\!\!\!\!\!\argmin_{V,W,A} && \!\!\!\!\!\!\!\!\!\!\!\!\!\Bigg[ \!\sum_{\obsum=1} \!\!\!\!\!\frac{\loss{\user,\movie}{\ytrue}{V^T \!W\!\!+\!\!A}}{\probobsum} + \lambda \!\left(||V||_F^2 \!\!+\!\!||W||_F^2\right)\!\Bigg] \label{eq:mf}
\end{eqnarray}
where $A$ encodes the offset terms and $\ypred^{ERM}=V^T W+A$.
Except for the propensities $\probobsum$ that act like weights for each loss term, the training objective is identical to the standard incomplete matrix factorization objective \cite{Koren,Steck,Hu/etal/08} with MSE (using Eq.~\eqref{eq:mse_loss}) or MAE (using Eq.~\eqref{eq:mae_loss}). 
So, we can readily draw upon existing optimization algorithms \citep[i.e.,][]{gemulla2011large,yu2012scalable} that can efficiently solve the training problem at scale. For the experiments reported in this paper, we use Limited-memory BFGS \cite{byrd1995limited}. Our implementation is available online\footnote{\href{https://www.cs.cornell.edu/~schnabts/mnar/}{https://www.cs.cornell.edu/$\sim$schnabts/mnar/}}.

Conventional incomplete matrix factorization is a special case of Eq.~\eqref{eq:mf} for MCAR (Missing Completely At Random) data, i.e., all propensities $\probobsum$ are equal. Solving this training objective for other $\lossum$ that are non-differentiable is more challenging, but possible avenues exist \cite{Joachims/05a,chapelle2010gradient}. Finally, note that other recommendation methods \citep[e.g.,][]{weimer2007maximum,lin2007projected} can in principle be adapted to propensity scoring as well. 

\section{Propensity Estimation for Observational Data} \label{sec:propest}
We now turn to the Observational Setting where propensities need to be estimated. 
One might be worried that we need to perfectly reconstruct all propensities for effective learning. However, as we will show, we merely need estimated propensities that are ``better'' than the naive assumption of observations being revealed uniformly, i.e., $\probobs = |\{(\user,\movie):\obsum=1\}|/\left(U\cdot I\right)$ for all users and items. The following characterizes ``better'' propensities in terms of the bias they induce and their effect on the variability of the learning process.
\begin{lemma}[Bias of IPS Estimator under Inaccurate Propensities]
\label{lemma:bias}
Let $\probobs$ be the marginal probabilities of observing an entry of the rating matrix $\ytrue$, and let $\estprobobs$ be the estimated propensities such that $\estprobobsum > 0$ for all $\user, \movie$. The bias of the IPS estimator Eq.~\eqref{eq:ips} using $\hat{P}$ is:
\begin{eqnarray}
\bias\left(\ipsestyhat\right) & \!\!\!=\!\!\! & \sum_{\user,\movie} \frac{\lossum}{\users \cdot \movies} \!\left[ 1 - \frac{\probobsum}{\estprobobsum} \right]. \label{eq:ips_bias} \:\:\:\:\:
\end{eqnarray}
\end{lemma}
In addition to bias, the following generalization error bound (proof in appendix) characterizes the overall impact of the estimated propensities on the learning process. 
\begin{theorem}[Propensity-Scored ERM Generalization Error Bound under Inaccurate Propensities]
\label{thm:prop_erm_bias}
For any finite hypothesis space of predictions $\hypspace=\{\ypred_1, ..., \ypred_{|\hypspace|}\}$, the transductive prediction error of the empirical risk minimizer $\ypred^{ERM}$, using the IPS estimator with estimated propensities $\hat{P}$ ($\estprobobsum > 0$) and given training observations $\obs$ from $\ytrue$ with independent Bernoulli propensities $\probobs$, is bounded by:
\vspace{-3mm}
\begin{eqnarray}
\risk{\ypred^{ERM}} &\!\!\!\le\!\!\!&  \ipsest{\ypred^{ERM}}{\estprobobs} + \frac{\Delta}{\users\!\cdot\!\movies}\sum_{\user,\movie} \left| 1 - \frac{\probobsum}{\estprobobsum} \right|  \nonumber\\
&& +\frac{\Delta}{\users\!\cdot\!\movies} \sqrt{\frac{\log\left(2 |\hypspace|/\eta\right)}{2}} \sqrt{\sum_{\user,\movie} \frac{1}{\estprobobsum^2}} \enspace .
\end{eqnarray}
\vspace{-3mm}
\end{theorem}
\vspace{-1mm}
The bound shows a bias-variance trade-off that does not occur in conventional ERM. In particular,
the bound suggests that it may be beneficial to overestimate small propensities, if this reduces the variability more than it increases the bias.

\subsection{Propensity Estimation Models.}
Recall that our goal is to estimate the probabilities $\probobsum$ with which ratings for user $\user$ and item $\movie$ will be observed. In general, the propensities
\vspace{-2mm}
\begin{equation}
    \probobsum = P(\obsum = 1\mid\xpropobs,\xprophid,\ytrue) \label{eq:propmod}
\end{equation}
can depend on some observable features $\xpropobs$ (e.g., the predicted rating displayed to the user), unobservable features $\xprophid$ (e.g., whether the item was recommended by a friend), and the ratings $\ytrue$. It is reasonable to assume that $\obsum$ is independent of the new predictions $\ypred$ (and therefore independent of $\lossum$) once the observable 
features are taken into account. The following outlines two simple propensity estimation methods, but there is a wide range of other techniques available \citep[e.g.,][]{Mccaffrey04propensityscore} that can cater to domain-specific needs.
\vspace{-3mm}
\paragraph{Propensity Estimation via Naive Bayes.}
\label{sec:prop_nb}
The first approach estimates $P(\obsum|\xpropobs,\xprophid,\ytrue)$ by assuming that dependencies between covariates $\xpropobs$, $\xprophid$ and other ratings are negligible. Eq.~\eqref{eq:propmod} then reduces to $P(\obsum|\ytrueum)$ similar to \citet{Marlin/Zemel/09}. We can treat $\ytrueum$ as observed, since we only need the propensities for observed entries to compute IPS and SNIPS. This yields the {\em Naive Bayes} propensity estimator:
\vspace{-2mm}
\begin{equation}
    P(\obsum = 1\mid\ytrueum = r) = \frac{P(\ytrue = r \mid\obs = 1) P(\obs = 1)}{P(\ytrue = r)}\enspace . \label{eq:propmodnb}
\end{equation}
We dropped the subscripts to reflect that parameters are tied across all $\user$ and $\movie$. 
Maximum likelihood estimates for $P(\ytrue = r \mid \obs=1)$ and $P(\obs = 1)$ can be obtained by counting
observed ratings in MNAR data.
However, to estimate $P(\ytrue = r)$, 
we need a small sample of MCAR data.
\vspace{-3mm}
\paragraph{Propensity Estimation via Logistic Regression}
\label{sec:prop_lr}
The second propensity estimation approach we explore 
(which does not require a sample of MCAR data) is based on logistic regression and is commonly used in causal inference \cite{Rosenbaum2002}. It also starts from Eq.~\eqref{eq:propmod}, but aims to find model parameters $\phi$ such that $\obs$ becomes independent of unobserved $\xprophid$ and $\ytrue$, i.e.,
$    P(\obsum|\xpropobs,\xprophid,\ytrue) = P(\obsum|\xpropobs,\phi).$
The main modeling assumption is that there exists a $\phi=(w, \beta, \gamma)$ such that 
$
    \probobsum = \sigma\left(w^T \xpropobsum + \beta_{\movie} + \gamma_{\user}\right).
$
Here, $\xpropobsum$ is a vector encoding all observable information about a user-item pair (e.g., user demographics, whether an item was promoted, etc.), and $\sigma(\cdot)$ is the sigmoid function. $\beta_{\movie}$ and $\gamma_{\user}$ are per-item and per-user offsets.

\section{Empirical Evaluation} \label{sec:exp}
We conduct semi-synthetic experiments to explore the empirical performance and robustness of the proposed methods in both the experimental and the observational setting. Furthermore, we compare against the state-of-the-art joint-likelihood method for MNAR data \cite{HL} on real-world datasets.

\subsection{Experiment Setup}
In all experiments, we perform model selection for the regularization parameter $\lambda$ and/or the rank of the factorization $d$ via cross-validation as follows. We randomly split the observed MNAR ratings into $k$ folds ($k = 4$ in all experiments), training on $k-1$ and evaluating on the remaining one using the IPS estimator. 
Reflecting this additional split 
requires scaling the propensities in the training folds by $\frac{k-1}{k}$ and those in the validation fold by $\frac{1}{k}$. 
The parameters with the best validation set performance are then used to retrain on all MNAR data.
We finally report performance on the MCAR test set for the real-world datasets, or using Eq.~\eqref{eq:risk} for our semi-synthetic dataset.

\subsection{How does sampling bias severity affect evaluation?} \label{sec:expbiaseval}
\begin{figure}[t]
\vspace*{-2mm}
\centering
\mbox{\hspace{-2mm}\includegraphics[width=0.53\linewidth,trim=0 0 15 0,clip]{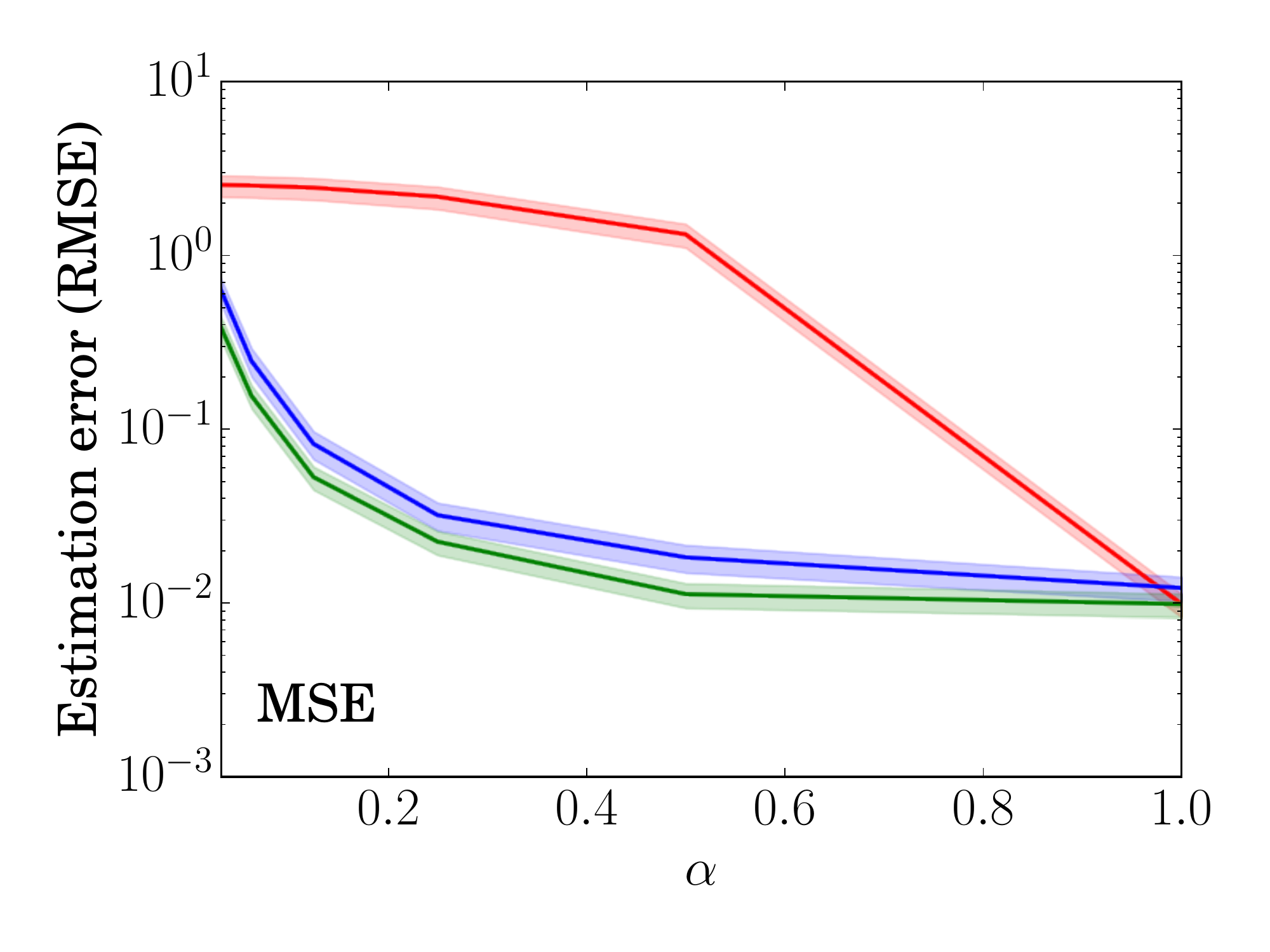}\hspace{-2mm}}
\mbox{\includegraphics[width=0.53\linewidth,trim=15 0 0 0,clip]{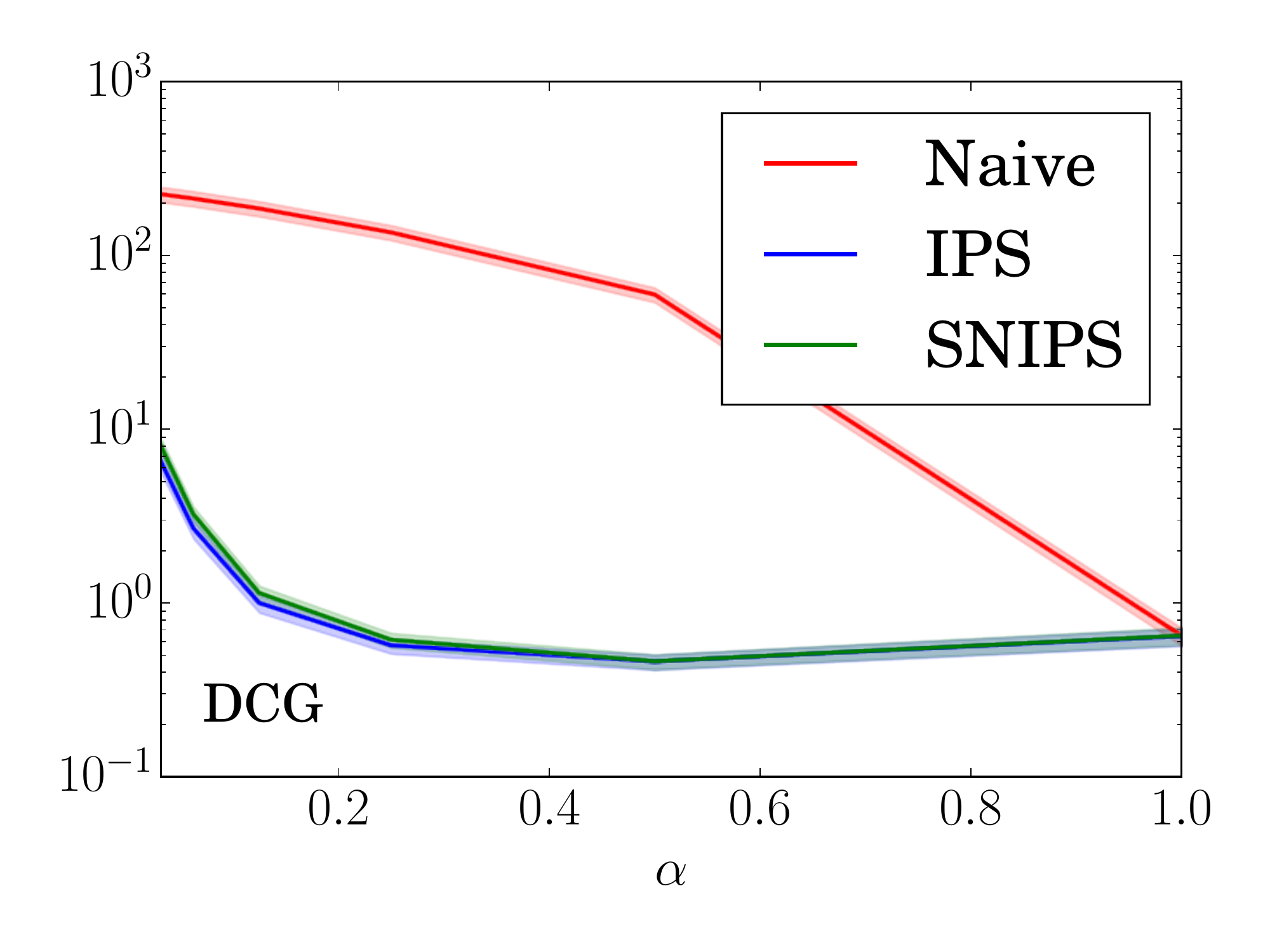}\hspace{-4mm}}
\vspace*{-5mm}
\caption{RMSE of the estimators in the experimental setting as the observed ratings exhibit varying degrees of selection bias. 
\label{fig:plot1}}
\end{figure}

First, we evaluate how different observation models impact the accuracy of performance estimates.
We compare the \naive\ estimator of Eq.~\eqref{eq:naiveest} for MSE, MAE and DCG with their propensity-weighted analogues, \ips\ using Eq.~\eqref{eq:ips} and \snips\  using Eq.~\eqref{eq:snips} respectively. Since this experiment requires experimental control of sampling bias, we created a semi-synthetic dataset and observation model.
\vspace{-3mm}
\paragraph{ML100K Dataset.} The ML100K dataset\footnote{\href{http://grouplens.org/datasets/movielens/}{http://grouplens.org/datasets/movielens/}} provides 100K MNAR ratings for 1683 movies by 944 users. To allow ground-truth evaluation against a fully known rating matrix, we complete these partial ratings using standard matrix factorization.
The completed matrix, however, gives unrealistically high ratings to almost all movies. We therefore adjust ratings for the final $\ytrue$ to match a more realistic rating distribution $\left[p_1, p_2, p_3, p_4, p_5 \right]$ for ratings 1 to 5 as given in \citet{Marlin/Zemel/09} as follows:  we assign the bottom $p_1$ fraction of the entries by value in the completed matrix a rating of 1, and the next $p_2$ fraction of entries by value a rating of 2, and so on.
Hyper-parameters (rank $d$ and L2 regularization $\lambda$) were chosen by using a 90-10 train-test split of the 100K ratings, and maximizing the 0/1 accuracy
of the completed matrix on the test set.
\vspace{-3mm}
\paragraph{ML100K Observation Model.} 
If the underlying rating is $4$ or $5$, the propensity for observing the rating is equal to $k$. For ratings $r < 4$, the corresponding propensity is $k \alpha^{4-r}$. For each $\alpha$, $k$ is set so that the expected number of ratings we observe is $5\%$ of the entire matrix.
By varying $\alpha > 0$, we vary the MNAR effect: $\alpha = 1$ is missing uniformly at random (MCAR), while $\alpha \rightarrow 0$ only reveals $4$ and $5$ rated items.
Note that $\alpha = 0.25$ gives a marginal distribution of observed ratings that reasonably matches the observed MNAR rating marginals on ML100K ($\left[ 0.06, 0.11, 0.27, 0.35,  0.21 \right]$ in the real data vs. $\left[ 0.06,  0.10,  0.25,  0.42,  0.17 \right]$ in our model).
\vspace{-3mm}
\paragraph{Results.} Table~\ref{tab:illuexp}, described in Section~\ref{sec:evalillu}, shows the estimated MAE and DCG@50 when $\alpha = 0.25$. Next, we vary the severity of the sampling bias by changing $\alpha \in ( 0,1 ]$. Figure~\ref{fig:plot1} reports how accurately (in terms of root mean squared estimation error (RMSE)) each estimator predicts the true MSE and DCG respectively.  These results are for the Experimental Setting where propensities are known. They are averages over the five prediction matrices $\ypred_i$ given in Section~\ref{sec:evalillu} and across $50$ trials. Shaded regions indicate a $95\%$ confidence interval.

Over most of the range of $\alpha$, in particular for the realistic value of $\alpha=0.25$, the \ips\ and \snips\ estimators are orders-of-magnitude more accurate than the \naive\ estimator. Even for severely low choices of $\alpha$, the gain due to bias reduction of \ips\ and \snips\ still outweighs the added variability compared to \naive. When $\alpha = 1$ (MCAR), \snips\ is algebraically equivalent to \naive, while \ips\ pays a small penalty due to increased variability from propensity weighting. For MSE, \snips\ consistently reduces estimation error over \ips\, while both are tied for DCG.

\subsection{How does sampling bias severity affect learning?} \label{sec:expbiaslearn}
\begin{figure}[t]
\vspace*{-3mm}
\centering
\mbox{\hspace{-2mm}\includegraphics[width=0.53\linewidth,trim=10 0 15 0,clip]{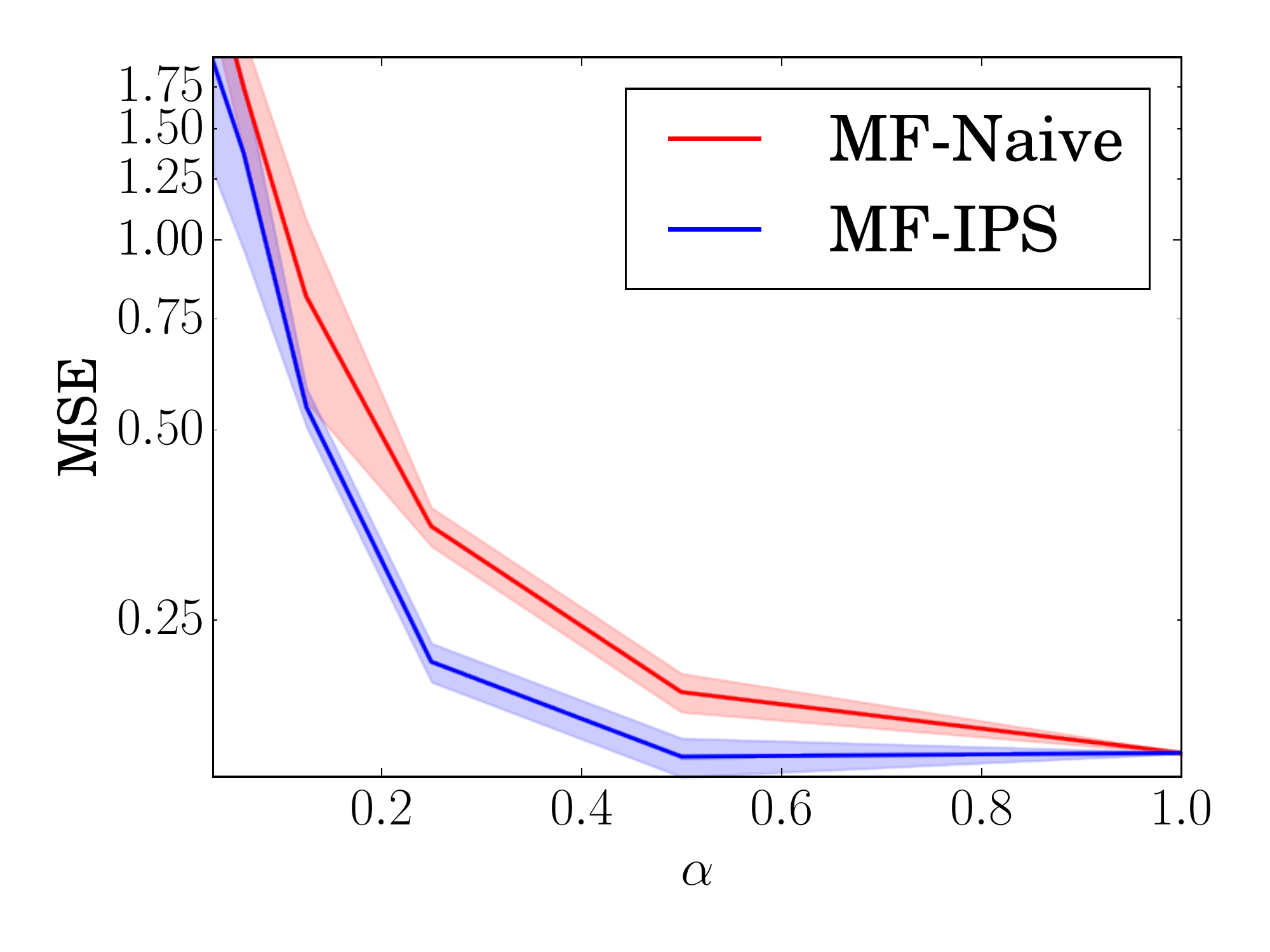}\hspace{-2mm}}
\mbox{\includegraphics[width=0.53\linewidth,trim=15 0 0 0,clip]{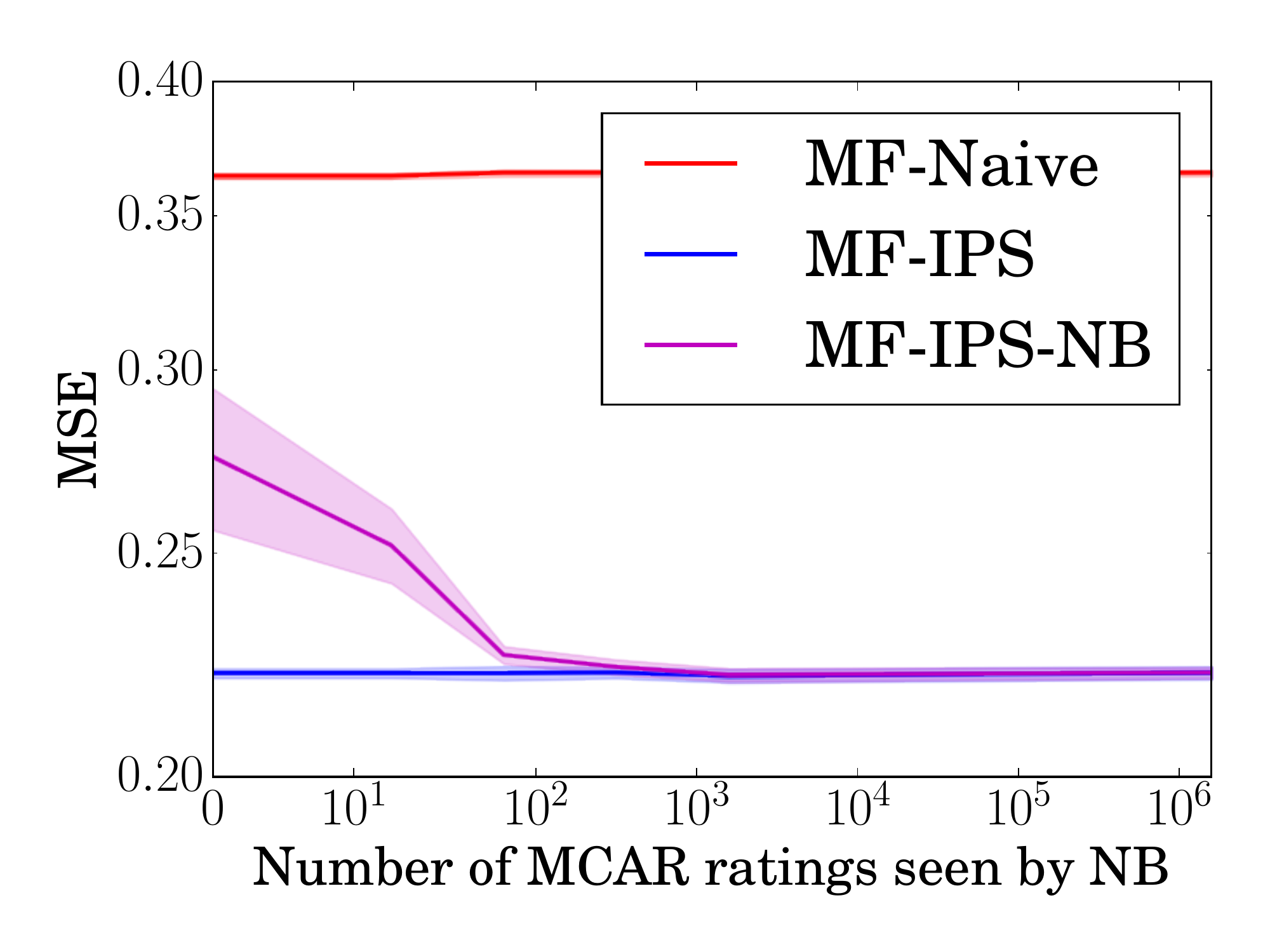}\hspace{-4mm}}
\vspace*{-5mm}
\caption{Prediction error (MSE) of matrix factorization methods as the observed ratings exhibit varying degrees of selection bias (left) and as propensity estimation quality degrades (right). \label{fig:plot2}}
\vspace{-1.5em}
\end{figure}

Now we explore whether these gains in risk estimation accuracy translate into improved learning via ERM, again in the Experimental Setting. Using the same semi-synthetic ML100K dataset and observation model as above, we compare our matrix factorization \ipsmf\ with the traditional unweighted matrix factorization \naivemf. Both methods use the same factorization model with separate $\lambda$ selected via cross-validation and $d=20$. The results are plotted in Figure~\ref{fig:plot2} (left), where shaded regions indicate $95\%$ confidence intervals over $30$ trials. The propensity-weighted matrix factorization $\ipsmf$\ consistently outperforms conventional matrix factorization
in terms of MSE. We also conducted experiments for MAE, with similar results.

\subsection{How robust is evaluation and learning to inaccurately learned propensities?} \label{sec:explearnedprop}
We now switch from the Experimental Setting to the Observational Setting, where propensities need to be estimated. To explore robustness to propensity estimates of varying accuracy, we use the ML100K data and observation model with $\alpha=0.25$. To generate increasingly bad propensity estimates, we use the Naive Bayes model from Section~\ref{sec:prop_nb}, but vary the size of the MCAR sample for estimating the marginal ratings $P(\ytrue = r)$ via the Laplace estimator. 

\begin{figure}[t]
\vspace*{-2mm}
\centering
\mbox{\hspace{-2mm}\includegraphics[width=0.53\linewidth,trim=0 0 15 0,clip]{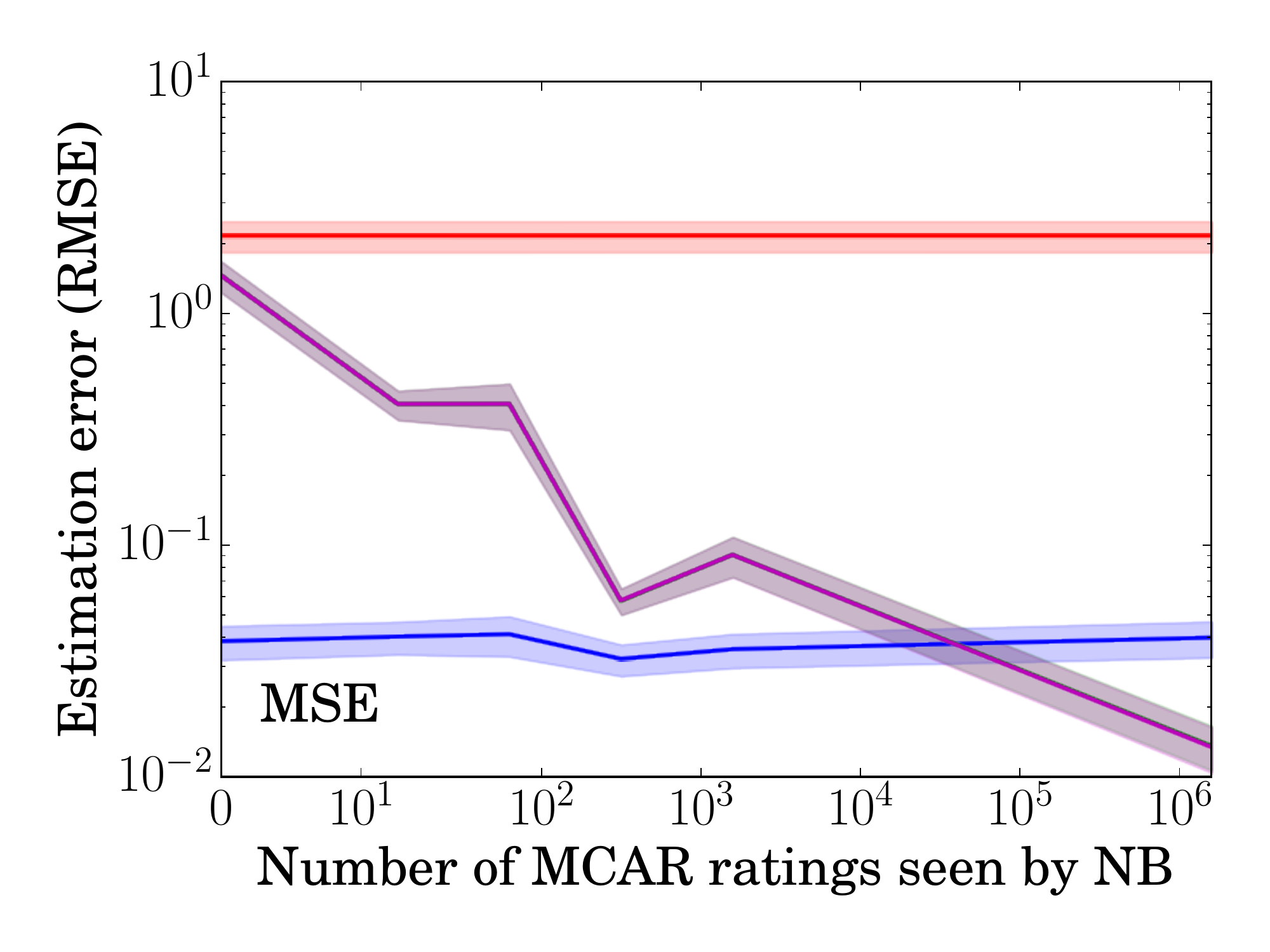}\hspace{-1mm}}
\mbox{\includegraphics[width=0.53\linewidth,trim=15 0 0 0,clip]{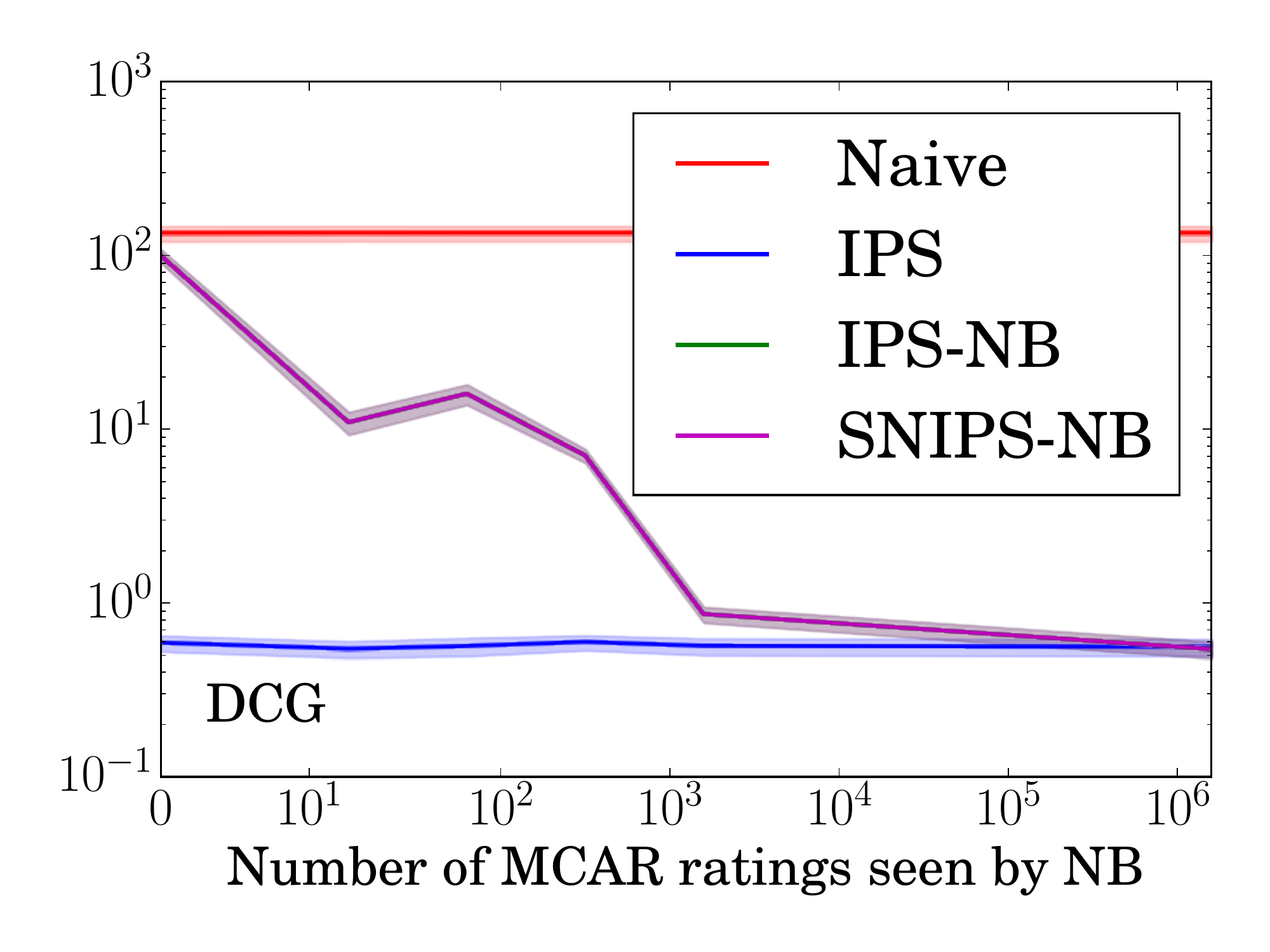}\hspace{-4mm}}
\vspace*{-5mm}
\caption{RMSE of \ips\ and \snips\ as propensity estimates degrade. \ips\ with true propensities and \naive\ are given as reference.
\label{fig:plot3}}
\vspace*{-5mm}
\end{figure}

Figure~\ref{fig:plot3} shows how the quality of the propensity estimates impacts evaluation using the same setup as in Section~\ref{sec:expbiaseval}. Under no condition do the \ips\ and \snips\ estimator perform worse than \naive. Interestingly, \ipsnb\ with estimated propensities can perform even better than \ipsgold\ with known propensities, as can be seen for MSE. This is a known effect, partly because the estimated propensities can provide an effect akin to stratification \cite{Hirano/etal/03,Wooldridge/07}.

Figure~\ref{fig:plot2} (right) shows how learning performance is affected by inaccurate propensities using the same setup as in Section~\ref{sec:expbiaslearn}. We compare the MSE prediction error of \ipsmfnb\ with estimated propensities to that of \naivemf\ and \ipsmf\ with known propensities. The shaded area shows the $95\%$ confidence interval over $30$ trials. Again, we see that \ipsmfnb\ outperforms \naivemf\ even for severely degraded propensity estimates, demonstrating the robustness of the approach.

\subsection{Performance on Real-World Data} \label{sec:exprealworld}
Our final experiment studies performance on real-world datasets. We use the following two datasets, which both have a separate test set where users were asked to rate a uniformly drawn sample of items.
\vspace{-3mm}
\paragraph{Yahoo! R3 Dataset.} 
This dataset\footnote{\href{http://webscope.sandbox.yahoo.com/}{http://webscope.sandbox.yahoo.com/}} \cite{Marlin/Zemel/09} contains user-song ratings. The MNAR training set provides over 300K ratings for songs that were self-selected by 15400 users. The test set contains ratings by a subset of 5400 users who were asked to rate 10 randomly chosen songs. 
For this data, we estimate propensities via Naive Bayes. As a MCAR sample for eliciting the marginal rating distribution, we set aside $5\%$ of the test set and only report results on the remaining $95\%$ of the test set.
\vspace{-3mm}
\paragraph{Coat Shopping Dataset.} 
We collected a new dataset\footnote{\href{https://www.cs.cornell.edu/~schnabts/mnar/}{https://www.cs.cornell.edu/$\sim$schnabts/mnar/}} simulating MNAR data of customers shopping for a coat in an online store. The training data was generated by giving Amazon Mechanical Turkers a simple web-shop interface with facets and paging. They were asked to find the coat in the store that they wanted to buy the most. Afterwards, they had to rate 24 of the coats they explored (self-selected) and 16 randomly picked ones on a five-point scale. The dataset contains ratings from $290$ Turkers on an inventory of $300$ items. The self-selected ratings are the training set and the uniformly selected ratings are the test set.
We learn propensities via logistic regression based on user covariates (gender, age group, location, and fashion-awareness) and item covariates (gender, coat type, color, and was it promoted). A standard regularized logistic regression \cite{scikit-learn}
was trained using all pairs of user and item covariates as features and cross-validated to optimize log-likelihood of the  self-selected observations.

\begin{table}[tb]
  \centering
  \scalebox{1.0}{
    \begin{tabular}{rrrrrr}
\toprule
          & \multicolumn{2}{c}{YAHOO}    &  & \multicolumn{2}{c}{COAT} \\
    \cmidrule{2-3} \cmidrule{5-6}
          & MAE  &  MSE &  \phantom{.}     & MAE  &  MSE \\
          \midrule
    \ipsmf   & \bf 0.810 & \bf 0.989 & & \bf 0.860 & \bf 1.093 \\
    \naivemf & 1.154 & 1.891 & & 0.920 & 1.202 \\
    HL MNAR & 1.177 & 2.175 & & 0.884 & 1.214 \\
    HL MAR   & 1.179 & 2.166 & & 0.892 & 1.220 \\
    \bottomrule
    \end{tabular}%
    }
    \caption{Test set MAE and MSE on the Yahoo and Coat datasets. \label{tab:realworld}}
    \vspace{-2mm}
\end{table}%

\vspace{-3mm}
\paragraph{Results.}
Table~\ref{tab:realworld} shows that our propensity-scored matrix factorization \ipsmf\ with learnt propensities substantially and significantly outperforms the conventional matrix factorization approach, as well as the Bayesian imputation models from \cite{HL}, abbreviated as HL-MNAR and HL-MAR (paired t-test, $p<0.001$ for all). This holds for both MAE and MSE. Furthermore, the performance of \ipsmf\ beats the best published results for Yahoo in terms of MSE (1.115) and is close in terms of MAE (0.770) (the CTP-v model of \cite{Marlin/Zemel/09} as reported in the supplementary material of \citet{HL}). For \ipsmf\ and \naivemf\ all hyperparameters (i.e., $\lambda \in \{10^{-6}, ..., 1\}$ and $d \in \{5,10,20,40\}$) were chosen by cross-validation. For the HL baselines, we explored $d \in \{5,10,20,40\}$  using software provided by the authors\footnote{\href{https://bitbucket.org/jmh233/missingdataicml2014}{https://bitbucket.org/jmh233/missingdataicml2014}} and report the best performance on the test set for efficiency reasons. Note that our performance numbers for HL on Yahoo closely match the values reported in \cite{HL}.

Compared to the complex generative HL models, we conclude that our discriminative \ipsmf\ performs robustly and efficiently on real-world data. We conjecture that this strength is a result of not requiring any generative assumptions about the validity of the rating model. Furthermore, note that there are several promising directions for further improving performance, like propensity clipping \cite{Strehl/etal/10}, doubly-robust estimation \cite{dudik11doublyrobust}, and the use of improved methods for propensity estimation \cite{Mccaffrey04propensityscore}.

\section{Conclusions}

We proposed an effective and robust approach to handle selection bias in the evaluation and training of recommender systems based on propensity scoring.
The approach is a discriminative alternative to existing joint-likelihood methods which are generative.
It therefore inherits many of the advantages (e.g., efficiency, predictive performance, no need for latent variables, fewer modeling assumptions) of discriminative methods.
The modularity of the approach---separating the estimation of the assignment model from the rating model---also makes it very practical. In particular, 
any conditional probability estimation method can be plugged in as the propensity estimator, and we conjecture that many existing rating models can be retrofit with propensity weighting without sacrificing scalability.

\section*{Acknowledgments} 
This research was funded in part under NSF Awards IIS-1247637, IIS-1217686, and IIS-1513692, and a gift from Bloomberg.


\end{document}